\pdfoutput=1

\documentclass[11pt]{article}

\usepackage[final]{acl}

\usepackage{times}
\usepackage{latexsym}

\usepackage[T1]{fontenc}

\usepackage[utf8]{inputenc}

\usepackage{microtype}

\usepackage{inconsolata}

\usepackage{graphicx}

\usepackage{algorithm}
\usepackage{algorithmic}
\usepackage{multirow}
\usepackage{xcolor}
\usepackage{fancyvrb}
\usepackage{listings}
\usepackage{booktabs}
\usepackage{amsmath}
\usepackage{todonotes}

\newcommand{\red}[1]{{\color{red}}}

\title{LineRetriever: Planning-Aware Observation Reduction for Web Agents}

\author{
Imene Kerboua\thanks{Equal contribution}\textsuperscript{1,2,7},
Sahar Omidi Shayegan\footnotemark[1]\textsuperscript{3,4,5}, \\
 Megh Thakkar\textsuperscript{3},
 Xing Han Lù\textsuperscript{4,5},
Massimo Caccia\textsuperscript{3}, \\
Véronique Eglin\thanks{Imene's Affiliated Supervisors}\textsuperscript{1,7},
Alexandre Aussem\footnotemark[2]\textsuperscript{6,7},
Jérémy Espinas\footnotemark[2]\textsuperscript{2},
Alexandre Lacoste\textsuperscript{3}
\\
\\
 \textsuperscript{1} INSA Lyon,
 \textsuperscript{2} Esker,
 \textsuperscript{3} ServiceNow Research,
 \textsuperscript{4} Mila AI Institute,
 \textsuperscript{5} McGill University,\\
 \textsuperscript{6} Université Claude Bernard Lyon 1,
 \textsuperscript{7} LIRIS
\\
 \small{
   \textbf{Correspondence:} \href{mailto:imene.kerboua@insa-lyon.fr}{imene.kerboua@insa-lyon.fr}
 }
}

\begin{document}

\maketitle
\begin{abstract}
While large language models have demonstrated impressive capabilities in web navigation tasks, the extensive context of web pages, often represented as DOM or Accessibility Tree (AxTree) structures, frequently exceeds model context limits. 
Current approaches like bottom-up truncation or embedding-based retrieval lose critical information about page state and action history. 
This is particularly problematic for adaptive planning in web agents, where understanding the current state is essential for determining future actions.
We hypothesize that embedding models lack sufficient capacity to capture plan-relevant information, especially when retrieving content that supports future action prediction. This raises a fundamental question: how can retrieval methods be optimized for adaptive planning in web navigation tasks?
In response, we introduce \textit{LineRetriever}, a novel approach that leverages a language model to identify and retrieve observation lines most relevant to future navigation steps. 
Unlike traditional retrieval methods that focus solely on semantic similarity, \textit{LineRetriever} explicitly considers the planning horizon, prioritizing elements that contribute to action prediction. 
Our experiments demonstrate that \textit{LineRetriever} can reduce the size of the observation at each step for the web agent while maintaining consistent performance within the context limitations. 
\end{abstract}

\section{Introduction}

Web agents powered by large language models (LLMs) face a critical challenge when processing modern websites: Web pages representations are often very long and exceed the context window limitations of even advanced LLMs. 
This constraint undermines web agents' effectiveness when crucial navigational information becomes unavailable during decision-making.

Information retrieval is an established field in Natural Language Processing and has become increasingly important in the context of LLMs due to their limited context windows. In tasks involving long or complex observations retrieval mechanisms help reduce input length while preserving task-relevant information. This focused context allows LLMs to reason more effectively, reducing errors caused by irrelevant or noisy inputs and enabling the model to concentrate on the most salient elements for decision-making. For example, retrieval-augmented generation (RAG) has been shown to improve factual accuracy by injecting relevant documents parts into the generation process \cite{lewis_retrieval-augmented_2021}.
As such, retrieval not only supports scalability but also enhances the accuracy and efficiency of LLM-driven agents.

In the web agents domain, prior research has employed retrieval mechanisms as a strategy for context reduction in observations. For example, \citet{dengMind2WebGeneralistAgent2023} uses a reranking embedding models, that given chunks of the DOM, ranks them from top-relevant to less relevant according to the current state fo the page and the task goal. \citet{luWebLINXRealWorldWebsite2024} uses a similar approach, only this time replacing the reranker which is computation-heavy with a lighter approach, a retrieval embedding model, this model is trained and expected to return the top-k relevant chunks from the DOM.

However, these approaches present limitations in a zero-shot setting. Relying on semantic similarity solely does not always provide all information needed by a working agent, because the observation stream in such environments not only contains information about the current state but also encapsulates the effects of previous actions on the interface, which helps define future actions.

Alternatively, some researchers have implemented a simple approach that truncates observations from the bottom to accommodate context-length constraints  \cite{drouinWorkArenaHowCapable2024, zhouWebArenaRealisticWeb2023}. 
Despite adequate performance on established benchmarks, there is no empirical evidence establishing a causal link between information loss from truncation and subsequent task failures.


In this work, we present \textit{LineRetriever}, a simple method that utilizes a smaller language model to select and extract observation lines with the highest relevance to subsequent navigation decisions. 
In contrast to conventional retrieval approaches that emphasize semantic relevance exclusively, \textit{LineRetriever} is asked to indirectly incorporate planning considerations. 

Additionally, we investigate whether smaller LLMs can effectively extract the most crucial information from the web page observation so that it can be used by a larger LLM serving as the action model. Specifically, we aim to use a small LLM to retrieve the subset of lines from the AxTree that are most relevant for achieving the task goal, given the current observation, the goal specification, and the history of actions taken by the agent.

Our experimental results demonstrate that \textit{LineRetriever} effectively minimizes observation size at each interaction step while sustaining comparable performance levels within established context boundaries. We list our contributions as follows: 
\begin{itemize}
    \item We introduce a simple yet novel method that reduces the observation size, creating more efficient web agents.
    \item We provide extensive experimental validation demonstrating \textit{LineRetriever}'s effectiveness across various web navigation tasks.
\end{itemize} 

\begin{figure}[t]
    \centering
    \includegraphics[width=1\linewidth]{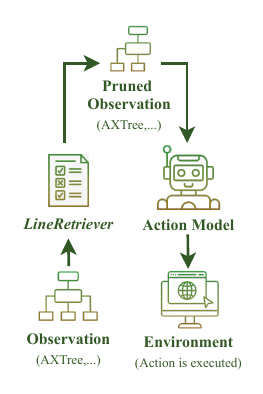}
    \caption{Overview of the \textit{LineRetrieverAgent} pipeline.}
    \label{fig:pipeline}
\end{figure}

\section{Related Work}

\subsection{Observation Processing in Web Agents}
The field of web agents has evolved rapidly in recent years, particularly with the integration of large language models (LLMs) for understanding and interacting with complex web interfaces \cite{nakanoWebGPTBrowserassistedQuestionanswering2022, zhouWebArenaRealisticWeb2023, drouinWorkArenaHowCapable2024}.
In general, approaches rely 3 types of observation: (1) AxTrees \cite{zhouWebArenaRealisticWeb2023, drouinWorkArenaHowCapable2024, sodhiStePStackedLLM2024, yangAgentOccamSimpleStrong2024}, (2) DOM \cite{luWebLINXRealWorldWebsite2024, dengMind2WebGeneralistAgent2023} or (3) screenshots \cite{ liuInstructionFollowingAgentsMultimodal2023, furutaMultimodalWebNavigation2023, yangSetMarkPromptingUnleashes2023, kohVisualWebArenaEvaluatingMultimodal2024}, each having their limitations. 
DOM-based approaches employ retrieval \cite{luWebLINXRealWorldWebsite2024} or reranking \cite{dengMind2WebGeneralistAgent2023} embedding models on DOM chunks, enabling agents to process only the most relevant information for task completion while filtering out noisy, irrelevant content that degrades performance. 
In contrast, AxTree-based methods have traditionally relied less on retrieval since AxTrees are typically more concise and contain fewer technical keywords than DOM representations, allowing them to fit within model context limits \cite{drouinWorkArenaHowCapable2024}.

However, as web applications become more complex and AxTrees grow larger, context length limitations and increasing costs due to longer pages processing are more frequent, necessitating intelligent filtering approaches.
Retrieval applied on the observation requires understanding the interactive elements and their relationships to user goals, making traditional embedding approaches less effective for navigation tasks where planning considerations and goal alignment are crucial. 

Our approach addresses these limitations by introducing an LLM-based retriever that explicitly incorporates planning context and user goal when filtering observations, enabling more effective selection of navigation-relevant content.

\subsection{Retrieval Methods for LLMs}

Early RAG approaches combined dense vector retrievers with generators to answer knowledge-intensive questions using retrieved evidence chunks \cite{lewis_retrieval-augmented_2021}, with extensions like REALM integrating retrieval during both pretraining and downstream tasks \cite{guu_realm_2020}. However, these embedding-based methods may not capture some context dependencies critical for planning in web navigation tasks.

Recent work demonstrates that LLMs can serve as intelligent retrievers by directly scoring or ranking documents. RankGPT showed that prompting GPT-3.5/4 in pairwise format enables high-quality reranking without fine-tuning \cite{sun_is_2024}, leading to efficient open-source variants like RankVicuna and RankZephyr distilled from GPT outputs \cite{pradeep_rankvicuna_2023}, and RankLLaMA adapted through contrastive fine-tuning \cite{ma_fine-tuning_2023}. These approaches better capture nuanced semantic relevance than traditional embedding-based retrievers, particularly valuable for ranking DOM elements or instructions in web agents.

However, recent frameworks like ReAct \cite{yao_react_2023} integrate reasoning with tool use, enabling the model to alternate between thought and action steps, which is particularly well-suited for interactive settings like web agents. Similarly, Self-Ask prompts models to generate sub-questions and retrieve answers before final composition \cite{press_measuring_2023}, while Toolformer teaches LLMs to call APIs mid-generation for dynamic retrieval \cite{schick_toolformer_2023}. These agentic methods enable goal-driven, retrieval-aware systems that adaptively access relevant context, particularly suited for interactive web environments where \textit{LineRetriever}'s observation-level context selection aligns with this structured reasoning paradigm.

While these approaches demonstrate the value of dynamic retrieval and LLM-based ranking, they lack specialized mechanisms for selecting relevant context at the granular observation level that web agents require for sequential decision-making. \textit{LineRetriever} addresses this gap by enabling context selection specifically at the observation level in web tasks, combining the semantic understanding of LLM-based retrievers with the step-by-step reasoning structure needed for effective web navigation.

\section{LineRetriever Agent}



\begin{figure*}
    \centering
    \includegraphics[width=1\linewidth]{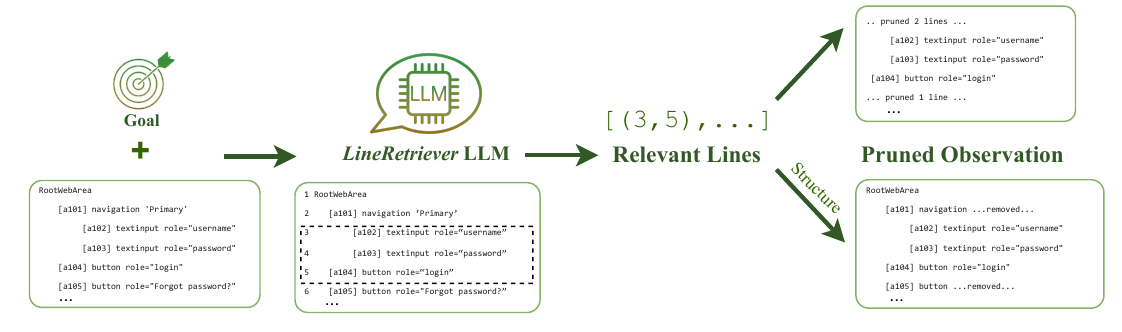}
    \caption{Example of LineRetriever functioning. This diagram shows how the LineRetriever LLM processes a user goal to identify and extract relevant lines from the AxTree for step completion. The system generates a list of line ranges, which are then used in postprocessing to filter out irrelevant lines, either by preserving the tree structure or by complete removal of non-essential content.}
    \label{fig:architecture}
\end{figure*}

\textit{LineRetriever} is a simple method designed to retrieve relevant information from observations to provide web agents with the information needed for effective action planning.
\textit{LineRetriever} is applied as a pre-processing method to each observation at each step of an episode.

Our approach utilizes a lightweight LLM as a selective filter. We construct a prompt containing three key components: (1) the current task goal, (2) the current observation with each line uniquely numbered for identification, and (3) optionally the complete interaction history documenting the agent's previous actions on the page.
The LLM analyzes this context to identify line ranges that are likely to contribute to future action decisions then selects relevant content directly.
Following the LLM's identification of relevant line ranges, post-processing filters the observation by retaining only the selected lines. We offer two approaches: direct line removal or structure-preserving filtering that maintains hierarchical relationships through parent element IDs and roles. The first approach delivers substantial compression while maintaining functional completeness, whereas the second method prioritizes structural integrity at the cost of reduced compression.
This streamlined observation is then passed to the agent, allowing it to operate within context constraints while retaining access to all task-critical information. Figure~\ref{fig:architecture} provides a visual overview of this process.





\section{Experimental Setup}
In this section, we provide details about the selected evaluation benchmarks (\ref{sec:benchmark}), relevant baselines (\ref{sec:baselines}), agents design (\ref{sec:agent_design}) and the evaluation metrics (\ref{sec:metrics}).

\subsection{Benchmarks}
\label{sec:benchmark}
To evaluate our agent we use 3 benchmarks: \textbf{(1) WorkArena L1} \cite{drouinWorkArenaHowCapable2024}, 
a real-world benchmark focused on routine knowledge work tasks. The main objective is to complete each task, given its goal and an accompanying web page, within a specified step limit
\textbf{(2) Weblinx} \cite{luWebLINXRealWorldWebsite2024}, a collection of real-world user tasks. This benchmark contains more complex and longer web pages, which require fitting the size of the observation into the LLM context window. \textbf{(3) WebArena} \cite{zhouWebArenaRealisticWeb2023}, a real-world tasks benchmark. To align with the evaluation setup from previous works, we use the BrowserGym test split \citep{dechezelles2025browsergymecosystemwebagent}, which defines $381$ tasks out of the full $812$ tasks available in WebArena. This splits enables fair comparison between zero-shot methods and models that were finetuned on a training subset of WebArena.

\subsection{Baselines}
\label{sec:baselines}
We identify two relevant baselines for comparison with our proposed approach.

\paragraph{Observation Bottom Truncation} 
We use GenericAgent \cite{drouinWorkArenaHowCapable2024}, an open-source generic agent available on the BrowserGym framework \cite{dechezelles2025browsergymecosystemwebagent}, which applies bottom-truncation for observations when they are too long. 
This agent as tested on multiple benchmarks and LLMs, which gives us a clear view of it's performance over different benchmarks.
See the work by \citet{drouinWorkArenaHowCapable2024} for more details on the truncation algorithm.

\paragraph{Embedding Retrieval} 
We build a baseline that leverages embeddings to retrieve relevant chunks.
Similarly to Dense Markup Ranker (DMR) method \cite{luWebLINXRealWorldWebsite2024}, we set the query to the task goal and history of previous interaction with the task. 
Chunks are built at each step based on the current observation, we set the chunk size to 100 tokens with an overlap of 10 tokens, we normalize embeddings and use \textit{cosine\_similarity} as a similarity measure. The final observation consists of up to the first 10 retrieved chunks, depending on availability. We use OpenAI ``text-embedding-3-small'' as the text embedding model.


\subsection{Agent Design}
\label{sec:agent_design}


We design our agents to operate under a standardized evaluation protocol across three benchmarks: WorkArena L1 (15 steps), Weblinx (single-step tasks), and WebArena (30 steps per task). Each agent is restricted to a maximum context length of 40,000 tokens except for the bottom-truncation agent with 10K.
The models we used were GPT 4.1 and GPT 4.1-mini.

\subsection{Metrics}
\label{sec:metrics}

\paragraph{Success Rate and Standard Error.}
For each agent and benchmark, we report the Success Rate (SR) with the Standard Error ($\pm$SE) over the benchmark.
We use BrowserGym and Agentlab \cite{dechezelles2025browsergymecosystemwebagent} frameworks to run our experiments as they unify the interface between agents and environments. 
We run \textbf{WorkArena L1} on 10 seeds for each task, which results in 330 tasks. 
\textbf{Weblinx} being a static dataset, the seed is set to 1, we evaluate agents on the \textit{test-iid} subset, which contains 2650 tasks. 
And for \textbf{WebArena} we run all tasks with 1 seed, which results in 381 tasks.

\paragraph{Observation Reduction Percentage.}
We quantify the reduction in observation size by comparing the retrieved observation ($o_r$) to the original observation ($o_i$) using the formula:
\[
\text{Reduction}(o_i) = 1 - \frac{|o_r|}{|o_i|}
\]
where $|o_i|$ and $|o_r|$ denote the lengths (e.g., token count) of the original and retrieved observations, respectively.

\newcommand{\SE}[1]{\scriptsize{\textcolor{gray}{$\pm$#1}}}
\newcommand{\EmptyCell}[0]{\textcolor{gray}{-}}

\begin{table*}[ht!]
    \centering
    \caption{Success Rates (SR) with Standard Deviation (\textcolor{gray}{$\pm$SE}) and Average Reduction (Avg. Reduc.) of the AxTree compared to the original for the baselines agents and our approach on WorkArena L1, Weblinx and WebArena benchmarks. EmbeddingRetrievalAgent and LineRetrieverAgent backbone model is GPT-4.1 for all agents. Due to budget constraints, two cells have been left empty (\EmptyCell).}
    \label{tab:baselines}
    \resizebox{\textwidth}{!}{
    \begin{tabular}{llllp{3em}lp{3em}lp{3em}}
         \toprule
          & & & \multicolumn{2}{l}{\textbf{WorkArena L1}} & \multicolumn{2}{l}{\textbf{Weblinx}} & \multicolumn{2}{l}{\textbf{WebArena}} \\
          \raisebox{-5ex}[0pt]{\textbf{Agent}} & \raisebox{-5ex}[0pt]{\textbf{Retriever Model}} & \raisebox{-5ex}[0pt]{\textbf{Pruning Strategy}} & \raisebox{-5ex}[0pt]{\textbf{SR (\%)}} & \textbf{Avg. Reduc. (\%)} & \raisebox{-5ex}[0pt]{\textbf{SR (\%)}} & \textbf{Avg. Reduc. (\%)} & \raisebox{-5ex}[0pt]{\textbf{SR (\%)}} & \textbf{Avg. Reduc. (\%)} \\
         \midrule
         GenericAgent-4.1 & N/A & Bottom-truncation & \textbf{52.7} \SE{2.7} & 0  & \underline{13.9} \SE{0.6} & 3 & \textbf{32.3} \SE{2.4} & 3 \\ 
         GenericAgent-4.1 & N/A & Bottom-truncation 10K & 49.1 \SE{2.8} & 16 & 13.1 \SE{0.6} & 18 & \EmptyCell & \EmptyCell \\   
         GenericAgent-4.1-mini & N/A & Bottom-truncation & 46.4 \SE{2.7} & 0 & 13.1 \SE{0.6} & 3 & 26.1 \SE{2.2} & 3 \\
         EmbeddingRetrievalAgent & Embedding & Chunk retrieval & 19.4 \SE{2.2} & 52 & 10.0 \SE{0.5} & 43 & 7.8 \SE{1.5} & 72 \\
         \midrule
         LineRetrieverAgent & 4.1-mini & LineRetriever & 44.8 \SE{2.7} & \textbf{61} & \textbf{14.1} \SE{0.6} & \textbf{72} & 24.9 \SE{2.2} & \textbf{73} \\
         LineRetrieverAgent & 4.1 & LineRetriever & 48.2 \SE{2.8}& \underline{58} & 13.9 \SE{0.6} & 75 & \EmptyCell & \EmptyCell \\
         \midrule
         LineRetrieverAgent & 4.1-mini & LineRetriever + Structure & \underline{49.1} \SE{2.8} & 30 & 13.7 \SE{0.6} & 18 & \underline{30.2} \SE{2.4} & 24 \\
         \bottomrule
    \end{tabular}
    }
\end{table*}

\section{Discussion}

In this section, we discuss the key insights from our experimental evaluation of \textit{LineRetriever}, examining the effectiveness of LLM-based retrieval compared to the embedding-based approach, the trade-offs between model size and structural augmentation, and the impact of observation reduction on web agent performance. We analyze how these findings inform the design of efficient web agents leveraging observation retrieval.

\paragraph{Embedding vs LLM Retrieval}
The embedding-based approach (EmbeddingRetrievalAgent) provides an interesting observation, failing entirely on all benchmarks compared to LLM-based retrieval (\textit{LineRetriever}), achieving 19.4\% success on WorkArena L1 in contrast to 49.1\% with \textit{LineRetriever}. This suggests that while embedding-based retrieval can capture semantic similarity, it may lack the contextual reasoning capabilities necessary for complex interactive tasks.

\paragraph{Retrieving Information with Small Models}
Based on the experimental results presented in Table \ref{tab:baselines}, the choice between large and small retriever models presents a nuanced trade-off between performance and computational efficiency. 
Our findings demonstrate that a small retriever model augmented with the structure of the observation achieve superior performance across benchmarks.
While without the structure, a bigger model is more relevant (\textit{LineRetriever}Agent with 4.1 and 4.1-min as retrievers achieve 48.2\% and 44.8\% respectively), although the performance drops slightly compared to the full observation processing (GenericAgent-4.1 achieves top results on 2 benchmarks out of 3). 
Importantly, these results suggest that while larger models typically provide enhanced retrieval capabilities, the structured representation of observations can compensate for model size limitations, enabling smaller models to achieve competitive or even superior performance when provided with appropriately organized contextual information.

\paragraph{Impact of Tree Reduction and Structure on Performance}
While \textit{LineRetriever} achieves substantial observation reduction of 61\% on WorkArena L1, 72\% on Weblinx, and 73\% on WebArena, this reduction is not without cost. As shown in Table \ref{tab:baselines}, GenericAgent-4.1, which retains the full bottom-truncated observation, still achieves the highest performance on WorkArena L1 and WebArena. LineRetrieverAgent, although more efficient, shows a modest drop in success rate, especially on tasks with more structured inputs.
This drop suggests that aggressively pruning the AxTree can disturb its structural and semantic coherence in ways that hurt downstream reasoning. One likely explanation is that the resulting pruned trees fall outside the distribution of AxTrees the model has implicitly learned to understand which leads to confusion or failure to ground interface elements. We observe that when structure is reintroduced (LineRetriever + Structure), performance improves again, supporting the idea that some hierarchical clues are essential.

\begin{figure}[ht]
    \centering
    \includegraphics[width=1\linewidth]{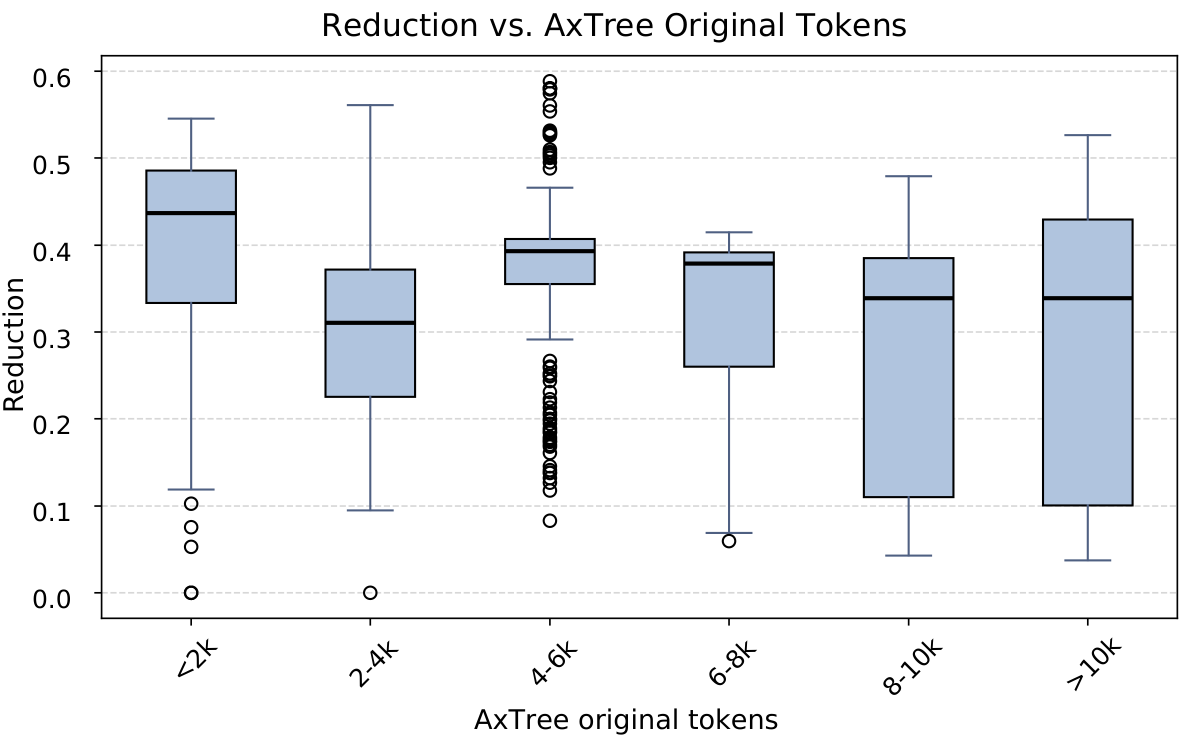}
    \caption{Box plot of token reduction versus AxTree original token count for the ``LineRetriever+Structure'' on WorkArena L1.}
    \label{fig:boxplots}
\end{figure}

Regrading the reduction ratios, Figure \ref{fig:boxplots} highlights that a higher token rate does not necessarily correlate with higher or lower reduction rate, suggesting that the reduction effectiveness depends more on the content of the observation rather than just the token count. 

In general, these results emphasize that observation reduction must not only aim to compress but also preserve the representational integrity of the input. The challenge is to remove irrelevant content without producing degenerate or overly abstracted AxTrees that break the model’s understanding of a web interface. Our ongoing work focuses on refining the structure-aware retrieval process to ensure the reduced observations remain familiar and navigable to the LLM policy.
\section{Conclusion}

We introduced LineRetriever, a planning-aware observation reduction method that uses a smaller language model to intelligently filter web page observations for web agents. Our approach achieves remakable observation reductions up to 73\% across multiple benchmarks while maintaining competitive performance. 
Key findings show that LLM-based retrieval on the observation significantly outperforms embedding-based approaches, and that preserving structural integrity is crucial for agent performance. 
While aggressive pruning can sometimes impact performance, LineRetriever demonstrates that planning-aware context reduction is both feasible and beneficial for scalable web agents. 


\bibliography{bibliography}

\appendix

\section{LineRetriever Prompt}


Figure \ref{fig:lr_prompt} shows \textit{LineRetriever} prompt.

\begin{figure*}
    \centering        
    \begin{lstlisting}[breaklines=true]
    SYSTEM:
    Your are part of a web agent who's job is to solve a task. Your are currently at a step of the whole episode, and your job is to extract the relevant information for solving the task. An agent will execute the task after you on the subset that you extracted. Make sure to extract sufficient information to be able to solve the task, but also remove informationcthat is irrelevant to reduce the size of the observation and all the distractions.

    USER:
    # Instructions:
    Extract the lines that may be relevant for the task at this step of completion. The subset should contain the relevant information to complete the task. Your answer should be a json list of indicating line numbers ranges e.g.: [(1,3), (20,25), (158,158), (200,250)]. Make sure to return information relevant to interact with the page.
    
    Answer format:
    <think>
    ...
    </think>
    <answer>
    ...
    </answer>

    # Goal:
    {goal}
    
    # History of interaction with the task:
    {history}
    
    # Observation:
    {axtree_txt}
    \end{lstlisting}
    \caption{\textit{LineRetriever} prompt.}
    \label{fig:lr_prompt}
\end{figure*}

\section{Cost Reduction with LLM Retrievers}

Let $\pi_{\theta_L}$ denote the agent's policy with parameters $\theta_L$ and $\pi_{\theta_S}$ denote the retrieval policy with parameters $\theta_S$, where $\theta_S \ll \theta_L$. For observation processing, we define $o_i$ as the original observation and $o_r$ as the reduced observation, with $|o_r| \leq \alpha \cdot |o_i|$ where $\alpha \in (0,1]$ represents the reduction ratio.

The cost comparison between our methods can be expressed as follows:
\begin{itemize}
    \item \textit{LineRetrieverAgent:} $C_S \cdot |o_i| + C_L \cdot |o_r|$, where $C_S$ is the cost of $\pi_{\theta_S}$
    \item \textit{GenericAgent:} $C_L \cdot |o_i|$, where $C_L$ is the cost of $\pi_{\theta_L}$.
\end{itemize}

For the \textit{LineRetrieverAgent} to be cost-effective, we require:

$$C_S \cdot |o_i| + C_L \cdot |o_r| \leq C_L \cdot |o_i|$$

Substituting $|o_r| = \alpha \cdot |o_i|$ and solving for $\alpha$:

$$C_S \cdot |o_i| + C_L \cdot \alpha \cdot |o_i| \leq C_L \cdot |o_i|$$
$$C_S + C_L \cdot \alpha \leq C_L$$
$$\alpha \leq \frac{C_L - C_S}{C_L}$$

In our experimental setting, $C_S = \text{0.4\$/1M tokens}$ and $C_L = \text{2\$/1M tokens}$. This yields:

$$\alpha \leq \frac{2 - 0.4}{2} \implies \alpha \leq 0.8$$

Therefore, cost efficiency is achieved when the observation size is reduced by at least 20\% ($1 - \alpha \geq 0.2$). 

\end{document}